\documentclass[a4paper, 11pt]{article}
\usepackage[english]{babel}
\usepackage{graphicx}
\usepackage{framed}
\usepackage[normalem]{ulem}
\usepackage{psfrag}
\usepackage{amsmath}
\usepackage{amsthm}
\usepackage{subfigure}
\usepackage{amssymb}
\usepackage{amsfonts}
\usepackage{enumerate}
\usepackage[utf8]{inputenc}
\usepackage[top=1.2 in,bottom=1.1in, left=1.2 in, right=1.2 in]{geometry}
\usepackage{algorithm,algorithmic}
\usepackage{framed} 
\usepackage{mdframed}
\usepackage{tcolorbox}
\usepackage{mathtools}
\usepackage{url}
\usepackage{booktabs}
\usepackage{amssymb,bm,mathrsfs}
\usepackage{authblk}
\usepackage[misc]{ifsym}
\allowdisplaybreaks
\numberwithin{equation}{section}

\theoremstyle{definition}
\newtheorem{theorem}{Theorem}
\newtheorem{lemma}{Lemma}

\newtheorem*{definition}{Definition}

\title{Deep Generative Learning via Variational  Gradient Flow}

\author[1]{Yuan Gao}
\author[2]{Yuling Jiao\thanks{
Yuling Jiao (yulingjiaomath@whu.edu.cn)}}
\author[3]{Yang Wang}
\author[1]{Yao Wang}
\author[3]{Can Yang\thanks{
Can Yang (macyang@ust.hk)}}
\author[3]{Shunkang Zhang}

\affil[1]{School of Mathematics and Statistics, Xi’an Jiaotong University}
\affil[2]{School of Statistics and Mathematics, Zhongnan University of Economics and Law}
\affil[3]{Department of Mathematics, The Hong Kong University of Science and Technology}

\date{January 24, 2019}

\begin{document}

\maketitle

\begin{abstract}
We propose a general framework to learn deep generative models via \textbf{V}ariational \textbf{Gr}adient Fl\textbf{ow} (VGrow) on probability spaces. The evolving distribution that asymptotically converges to the target distribution is governed by a vector field, which is the negative gradient of the first variation of the $f$-divergence between them. We prove that the evolving distribution coincides with the pushforward distribution through the infinitesimal time composition of residual maps that are perturbations of the identity map along the vector field. The vector field depends on the density ratio of the pushforward distribution and the target distribution, which can be consistently learned from a binary classification problem. Connections of our proposed VGrow method with other popular methods, such as VAE, GAN and flow-based methods, have been established in this framework, gaining new insights of deep generative learning. We also evaluated several commonly used divergences, including Kullback-Leibler, Jensen-Shannon, Jeffrey divergences as well as our newly discovered ``logD'' divergence which serves as the objective function of the logD-trick GAN. Experimental results on benchmark datasets demonstrate that VGrow can generate high-fidelity images in a stable and efficient manner, achieving competitive  performance with state-of-the-art GANs.
\end{abstract}

\section{Introduction}\label{sec1}

Learning the generative model, i.e., the underlying data generating distribution,
based on large amounts of data is one the  fundamental task  in machine
learning and statistics \cite{salakhutdinov2015learning}. Recent advances in deep generative models  have provided novel techniques for unsupervised and semi-supervised learning, with broad application varying from image synthesis \cite{reed16}, semantic image editing \cite{zhu16}, image-to-image translation \cite{zhu17} to low-level image processing \cite{ledig17}.  Implicit  deep  generative model is a powerful and flexible framework to approximate the target  distribution by learning deep  samplers \cite{mohamed2016learning}
including \emph{Generative adversarial networks} (GAN) \cite{goodfellow14} and likelihood based models,  such as  \emph{variational auto-encoders} (VAE) \cite{kingma14}  and \emph{flow based methods} \cite{dinh2014nice}, as their main representatives.
The above mentioned implicit deep generative models  focus on learning a deterministic or stochastic  nonlinear mapping that can transform  low dimensional latent samples from referenced simple distribution to  samples that   closely match the target  distribution.

GANs build a minmax two player game between the  generator and discriminator.
 During the training, the generator transforms samples from a simple reference distribution into samples that would hopefully    to deceive the discriminator, while the discriminator conducts  a differential  two-sample test to distinguish the generated samples from the observed samples.
The objective of vanilla GANs amounts to the \emph{Jensen-Shannon} (JS) divergence between the learned distribution  and target  distributions. The vanilla  GAN generates sharp image samples but suffers form the instability issues \cite{arjovsky17}.
A myriad of extensions to vanilla GANs have been investigated, both theoretically or empirically, in order to achieve a stable training  and high quality   sample generation. Existing works  include but are not limited to
designing  new learning procedures or network architectures   \cite{denton15,radford15,zhang17,zhao17,arora17,tao18,brock18},
	and  seeking  alternative distribution discrepancy measures as loss  criteria in feature or data space \cite{li15, dziugaite15, li17, sutherland17, binkowski18,arjovsky17,mao17,mroueh17},
	and  exploiting  insightful regularization methods  \cite{che17,gulrajani17,miyato18,zhang18},
	and  building  hybrid models  \cite{donahue17, tolstikhin17, dumoulin17,ulyanov18,huang18}.

VAE approximately minimizes  the \emph{Kullback-Leibler} (KL) divergence between the transformed  distribution and the target distribution  via minimizing a surrogate loss  , i.e., the negative evidence lower bound  defined as the reconstruction loss  plus the regularization loss, where the reconstruction loss measures the difference between the decoder and the encoder, and the regularization loss measures the difference between the encoder and the simple latent prior distribution \cite{kingma14}.  VAE enjoys optimization stability but was disputed for generating blurry image samples caused by the  Gaussian decoder and the marginal log-likelihood based loss \cite{tolstikhin17}.
Adversarial auto-encoders  \cite{makhzani15} use GANs to  penalize  the discrepancy between the   aggregated posterior of latent codes and the simple prior distribution.  Wasserstein auto-encoders  \cite{tolstikhin18} that  extend  the adversarial auto-encoders  to general  penalized optimal transport  objectives \cite{bousquet17} alleviate  the blurry.  Similar ideas are found in some works on disentangled representations of natural images \cite{higgins16, kumar18}.

 Flow based methods minimize exactly  the negative log-likelihood, i.e., the KL divergence,   where the model density  is the pushforward density of simple reference density through a sequence of learnable  invertible transformations  called  normalizing  flow \cite{rezende2015variational}. The research of flow based generative models mainly focus on designing  the
  neural network  architectures  to trade off the   representative power and   the computation complexity  of the log-determinants \cite{dinh2014nice,dinh2016density,kingma2016improved,papamakarios2017masked,kingma2018glow}.

In this paper, we propose a general framework to learn a deep generative model   to sample from the target distribution   via combing the strengths of  variational  gradient flow  (VGrow) on  probability space, particle optimization and   deep  neural network.
Our method aims to find    a deterministic transportation
map that transforms low dimensional  samples from a simple reference distribution,  such as  Gaussian distribution or uniform distribution,
 into samples from underlying  target
distribution. The evolving distribution  that asymptotically
converges to the target distribution is governed by a vector field, which is   the negative  gradient of the   first variation  of   the $f$-divergence between the  the evolution distribution and the target distribution.
We prove that the evolution distribution coincides with the pushforward distribution through the infinitesimal time composition of residual maps that are perturbations of the identity map along the  vector field.
  At the population level, the vector field  only depends on the density ratio of the pushforward distribution and the target distribution, which can be  consistently learned from   a binary classification problem to distinguish the  observed data sampling from the target distribution
 from the generated  data sampling from pushforward distribution.
Both the transform and binary classifier   are parameterized  with deep convolutional  neural networks and trained via stochastic gradient descent (SGD).
 Connections of our proposed VGrow method with other popular methods, such as VAE, GAN and flow-based methods, have been established in this our framwork, gaining new insights of deep generative learning. We also evaluated several commonly used divergences, including Kullback-Leibler, Jensen-Shannon, Jeffrey divergences as well as our newly discovered ``logD'' divergence  serving  as the objective function of the logD-trick GAN,
 which is of independent interest of  its own.    We test VGrow with the above mentioned  four divergences  on four benchmark datasets including MNIST \cite{lecun98}, FashionMNIST \cite{xiao17}, CIFAR10 \cite{krizhevsky09} and CelebA \cite{liu15}. The VGrow learning procedure  is very stable, as indicted from our established theory. The resulting   deep  sampler can obtain realistic looking images, achieving  competitive performance   with state-of-the-art GANs.
 The code of VGrow is available at \url{https://github.com/xjtuygao/VGrow}.


\section{Background, Notation and Theory}\label{sec2}

Let $\{\mathbf{X}_i\}_{i=1}^N \subset \mathbb{R}^d$  be  independent and  identically distributed samples from an unknown target  distribution $\nu$ with density $p(\mathbf{x})$ with respective to
the Lebesgue measure (we made the same assumption for the distributions  in this paper). We aim  to learn the distribution $\nu$ via constructing variational gradient flow  on Borel probability $\mathcal{P}(\mathbb{R}^d)$. To this end, we need the following backround detail studied in   \cite{ambrosio2008gradient}.

Given  $\mu \in \mathcal{P}(\mathbb{R}^d)$  with density $q(\mathbf{x})$,  we use the $f$-divergence $\mathbb{D}_f(\mu | \nu)$
to measure the discrepancy between  $\mu$ and  $\nu$ which is defined as
\begin{equation}
\mathbb{D}_f(q(\mathbf{x}) | p(\mathbf{x})) = \int p(\mathbf{x}) f\left(\frac{q(\mathbf{x})}{p(\mathbf{x})} \right)d\mathbf{x},
\end{equation}
where $f: \mathbb{R}^1 \rightarrow \mathbb{R}^1$ is convex and $f(1) = 0$. We  use $\mathcal{F}(\cdot)$ to  denote the energy  functional
$ \mathbb{D}_f(\cdot|p(\mathbf{x})) :\mathcal{P}(\mathbb{R}^d) \rightarrow \mathcal{R}^{+}$ for simplicity. Obviously $\mathcal{F}(q) \geq 0$ and $\mathcal{F}(q) = 0 $ iff $q(\mathbf{x}) = p(\mathbf{x}).$
\begin{lemma}\label{lem1}
Let $\frac{\delta\mathcal{F}}{\delta{q}}(q): \mathcal{R}^{d}\rightarrow \mathcal{R}^{1}$ be the first variation of  $\mathcal{F}(\cdot)$ at $q(\mathbf{x})$.   $\left(\frac{\delta\mathcal{F}}{\delta{q}}(q)\right) (\mathbf{x})= f{'}(r(\mathbf{x}))$ with $r(\mathbf{x}) = \frac{q(\mathbf{x})}{p(\mathbf{x})}.$
\end{lemma}

Considering a curve $\mu_t : t\in \mathbb{R}^{+} \rightarrow \mathcal{P}(\mathbb{R}^d)$ with density $q_{t}(\mathbf{x})$.
Let $$\textbf{v}_t(\mathbf{x}) = - \nabla \left(\frac{\delta\mathcal{F}}{\delta q_t}(q_t)\right)(\mathbf{x}) : \mathcal{R}^{+}\rightarrow (\mathcal{R}^{d}\rightarrow \mathcal{R}^{d})$$ be the vector field and $r_t(\mathbf{x}) = \frac{q_t(\mathbf{x})}{p(\mathbf{x})}.$
\begin{definition}
We call $\mu_t$ is a variational gradient  flow   of the  energy functional $\mathcal{F}(\cdot)$  governed by the vector field $\textbf{v}_t(\mathbf{x})$
if satisfies the  Vlasov-Fokker-Planck equation
\begin{equation}\label{vfp}
\frac{d }{d t}q_t(\mathbf{x})
 = -\nabla\cdot(q_t(\mathbf{x})\textbf{v}_t(\mathbf{x}))).
\end{equation}
\end{definition}
As shown in the following Lemma \ref{lem2},  the energy functional  $\mathcal{F}(\cdot)$ is decreasing along the curve $\mu_t$. As a consequence,  the limit of $q_t(\mathbf{x})$ is the  target $p(\mathbf{x})$ as $t\rightarrow \infty$.
\begin{lemma}\label{lem2}
$$\frac{d }{d t} \mathcal{F}(q_t) = - \mathbb{E}_{X \sim q_t} [\|\textbf{v}_t(X)\|^2]$$
\end{lemma}


For any fixed time  $t\in\mathcal{R}^{+}$, let $X$ be a random variable with distribution  $q_t(\mathbf{x})$. Let  $\mathbf{h}(\mathbf{x}) : \mathcal{R}^{d} \rightarrow \mathcal{R}^{d}$ be an element of the Hilbert space $\mathcal{H}(q_t) = [L^2(q_t)]^{d}$ and  $s\in \mathcal{R}^{+}$ be a small positive number. Define a residual map $\mathbb{T}_{s, \mathbf{h}} :\mathcal{R}^{d} \rightarrow \mathcal{R}^{d}$ as a small permutation of identify map $\mathbf{id}$ along $\mathbf{h}(\mathbf{x})$,
i.e.,  $$\mathbb{T}_{s, \mathbf{h}} = \mathbf{id} + s \mathbf{h}.$$ Let $\mathbb{T} ^{-1}_{s, \mathbf{h}}$ be  the inverse of  $\mathbb{T}_{s, \mathbf{h}} $, which  is well defined when  $s$ is small enough. By change of variable formula,  the  density of pushforward distribution of  random variable    $ \mathbb{T}_{s, \mathbf{h}} (X) $  is  $$({\mathbb{T}_{s, \mathbf{h}}}_\#q_t) (\mathbf{x}) = q_t(\mathbb{T} ^{-1}_{s, \mathbf{h}}(\mathbf{x})) |\mathrm{det}(\nabla_\mathbf{x} \mathbb{T} ^{-1}_{s, \mathbf{h}}(\mathbf{x}))|.$$
Let $$\mathcal{L}(\mathbf{h}) = \mathbb{D}_f[({\mathbb{T}_{s, \mathbf{h}}}_\#q_t) (\mathbf{x}) | p(\mathbf{x})]$$ denote the  functional of $\mathbf{h}$ mapping from $\mathcal{H}(q_t) $ to $\mathbb{R}^1$. It is natural to find $\mathbf{h}$ satisfying  $\mathcal{L}(\mathbf{h})< \mathcal{L}(\mathbf{0}) $, which indicates the pushforward distribution $({\mathbb{T}_{s, \mathbf{h}}}_\#q_t) (\mathbf{x})$ is much closer to $p(\mathbf{x})$ than $q_t(\mathbf{x})$. We find such $\mathbf{h}$ via calculating the first
variation of the functional $\mathcal{L}(\mathbf{h})$ at $\mathbf{0}.$

\begin{theorem}\label{th1}
For any $\textbf{g} \in \mathcal{H}(q_t)$, if the vanishing condition $\lim \limits_{\|\textbf{x}\|\rightarrow \infty } \|f'(r_t(\mathbf{x}))q_t(\mathbf{x})\mathbf{g}(\mathbf{x})\| = 0$ is satisfied, then
$$\langle\frac{\delta\mathcal{L}}{\delta \mathbf{h}}(\mathbf{0}), \textbf{g}\rangle_{\mathcal{H}(q_t)} =   \langle f''(r_t)\nabla r_t, \textbf{g}\rangle_{\mathcal{H}(q_t)}.$$
\end{theorem}
The vanishing condition assumed in Theorem \ref{th1} holds when the densities   have compact supports or with light tails.
Theorem \ref{th1} shows that the  residual map  defined as  a small   perturbation of identity map along the vector field $\mathbf{v}_t(\mathbf{x})$ can push samples
from $q_t(\mathbf{x})$  into samples more likely  sampled from $p(\mathbf{x})$.

\begin{theorem}\label{th2}
The evolution  distribution of $q_t$ under infinitesimal pushforward  map  $\mathbb{T}_{s, \mathbf{v}_t}$   satisfies the Vlasov-Fokker-Planck equation \eqref{vfp}.
\end{theorem}

As consequences of Theorem \ref{th2}, we know the pushforward distribution through the  residual maps  with  infinitesimal time perturbations is
the same as the variational  gradient flow.  This connection motivates us to approximately solve the  Vlasov-Fokker-Planck equation \eqref{vfp} via finding a pushforward map defined as  composition of  sequences of discreet time  residual maps with  small stepsize  as long as   we can learn the vector field $\mathbf{v}_t(\mathbf{x})$.
By definition, the vector field $\mathbf{v}_t(\mathbf{x})$ is an explicit   function of density ratio  $r_t(\mathbf{x})$, which is  well studied, see for example, \cite{sugiyama2012density}.

\begin{lemma}\label{lem3}
Let $(X,Y)$ be random variable  pair samples from $ p(\mathbf{x},y)$  with binary $Y\sim $ marginal distribution $ p(y)$ taking value in  $ \{-1,+1\}$.
Denote  $q(\mathbf{x})=p(\mathbf{x}|Y=-1), ~p(\mathbf{x})=p(\mathbf{x}|Y=1)$ and $r(\mathbf{x}) = \frac{q(\mathbf{x})}{p(\mathbf{x})}. $
Let
$$d^*(\mathbf{x}) = \arg\min\limits_{d(\mathbf{x})}\mathbb{E}_{(X, Y) \sim p(\mathbf{x}, y)}\log(1+\exp(-d(\mathbf{X}) Y)).$$
If $p(Y=1)= p(Y=-1)$,
then $r(\mathbf{x}) = \exp^{-d^*(\mathbf{x})}.$
\end{lemma}

According to Lemma \ref{lem3}, we can estimate the density ratio $r_t(\mathbf{x}) = \frac{q_t(\mathbf{x})}{p(\mathbf{x})}$ via samples.  Let $Z_i, X_i, i =1,...,N$  be samples from  $q_t$ and $ p(\mathbf{x})$, respectively. We introduce a random variable $Y$, and
assign a label $Y_i = -1 $ for $Z_i$ and $Y_i =  1$ for $X_i$.
Define
\begin{equation}\label{lr}
\hat{d}(x) = \arg\min\limits_{d(\mathbf{x})} \sum_{i=1}^N (\log(1+\exp(-d(X_i))
+ \log(1+\exp(d(Z_i))).
 \end{equation}
  Then
  $\hat{r}(\mathbf{x}) = e^{-\hat{d}(\mathbf{x})}$  consistently estimates $r_t(\mathbf{x})$ as $N \rightarrow \infty.$

\section{Variational gradient flow (VGrow) learning procedure}\label{sec3}
  With data   $\{\mathbf{X}_i\}_{i=1}^N \subset \mathbb{R}^d$ samples from an unknown target  distribution  $p(\mathbf{x})$, our goal is to learn    a deterministic transportation
map  that transforms low dimensional  samples  from a simple reference distribution  such as a Gaussian distribution or a uniform distribution
 into samples  from underlying  target $p(\mathbf{x})$.

 To this end, we parameterize the sought transform via     a  deep neural network   $G_{\theta}:   \mathcal{R}^{\ell} \rightarrow \mathcal{R}^{d}$   with $\ell \ll d$, where   $\theta$ denotes its parameter. We sample particles $W_i$ from simple reference distribution and transform them  into $Z_i$  with the initial   $G_{\theta}$.  We do the following two steps iteratively. First,  we  learn a density ratio via solving \eqref{lr} with  real data $X_i$ and generated data  $Z_i$, where we  parameterize  $d(\cdot)$  into   a  neural network $D_{\phi}(\cdot)$. Then, we define residual map $\hat{\mathbb{T}}$ using  the estimated vector field with a small step size and update $Z_i$ by $\widehat{\mathbb{T}}(Z_i)$.
 According to the theory we discussed in Section \ref{sec3}, the above iteratively two steps can get particles  $Z_i$ more likely sampled from $p(\mathbf{x})$.
 So, we can update the generator $G_{\theta}$ via fitting the pairs $(W_i, Z_i).$
We can repeat the above whole procedure as desired with warmsart. We give the detail description of VGrow learning procedure as follows.
\begin{itemize}
\item Outer loop
  \begin{itemize}
   \item  Sample  $W_i \in \mathcal{R}^{\ell}, i =1,...,N$   from the simple reference distribution and let  particles $Z_i = G_{\theta}(W_i)$.\\
   Inner loop
      \begin{itemize}
       \item  Restrict    $d(\cdot)$ in  \eqref{lr} be  a  neural network $D_{\phi}(\cdot)$ with parameter $\phi$ and solve  \eqref{lr} with SGD  to get   $\hat{r}(\mathbf{x}) = e^{-D_{\phi}(\mathbf{x})}$.
       \item Define  the residual map $ \widehat{\mathbb{T}}  = \mathbf{id} + s \widehat{\mathbf{h}}$ with a small step size $s$, where $\widehat{\mathbf{h}}(\mathbf{x}) = - f{''}(\hat{r}(\mathbf{x}))\nabla \hat{r}(\mathbf{x})$.
        \item  Update the particles  $Z_i = \widehat{\mathbb{T}}(Z_i), i =1,...,N$.
       \end{itemize}
       End inner loop.
     \item  Update the parameter $\theta$ via solving $\min_{\theta}\sum_{i=1}^N \|G_{\theta}(W_i) - Z_i\|^2$  with SGD.
    \end{itemize}
\item End outer loop
\end{itemize}

We consider  four  divergences in our paper. The form of the four divergences and their second order derivatives are shown in Table \ref{div}.
They are  the three  commonly used divergences, including Kullback-Leibler (KL), Jensen-Shannon (JS), Jeffrey divergences, as well as our newly discovered ``logD'' divergence  serving  as the objective function of the logD-trick GAN, which  to the best of knowledge is a new result.

\begin{theorem}\label{th3}
At the  population level, the  logD-trick GAN \cite{goodfellow14} minimizes the ``logD'' divergence $\mathbb{D}_f(q(\mathbf{x}) | p(\mathbf{x}))$, with $f(u) = (u+1) \log (u+1)-2\log 2$, where
 $q(\mathbf{x})$ is the distribution of generated data.
\end{theorem}

\begin{table}[H]
\caption{Four representative $f$-divergences}
\label{div}
\vskip 0.15in
\begin{center}
\begin{small}
\begin{rm}
\begin{tabular}{lcc}
\toprule
$f$-Div		& $f(u)$ 								 & $ f''(u)$ \\
\midrule
KL    		& $u \log u$ 							 & $\frac{1}{u}$ 	  \\
JS 			& $-(u+1) \log \frac{u+1}{2} + u \log u$ & $\frac{1}{u(u+1)}$ \\
logD    	& $(u+1) \log (u+1) - 2 \log 2$ 		 & $\frac{1}{u+1}$ 	  \\
Jeffrey    	& $(u-1) \log u$ 						 & $\frac{u+1}{u^2}$  \\
\bottomrule
\end{tabular}
\end{rm}
\end{small}
\end{center}
\vskip -0.1in
\end{table}
\section{Related Works}

We discuss connections between our proposed VGrow learning procedure  and related works, such as
  VAE, GAN and flow-based methods.

VAE \cite{kingma14} is formulated  as maximizing a lower bound based on the  KL divergence. Flow based methods \cite{dinh2014nice,dinh2016density} minimize the KL divergence between target and a model, which is  pushforward density of a simple reference density through a sequence of learnable  invertible transformations.
The fow based methods parameterize these  transforms via special designed neural networks facilitating log determinant  computation   \cite{dinh2014nice,dinh2016density,kingma2016improved,papamakarios2017masked,kingma2018glow}
 and train it  using MLE.
 Our VGrow also learns a sequence of simple residual maps guided form the variational gradient flow in probability space,  which is  quite different from  the flow based method in principle.

The original vanilla GAN  and the logD-trick GAN  \cite{goodfellow14} minimize the JS divergence and the ``logD'' divergence, respectively, as shown in Theorem \ref{th3}.
This idea can be extended  to a general $f$-GAN \cite{nowozin16}, where the general $f$-divergence is used. However, the GANs based on $f$-divergence are formulated
to solve the dual problem. In contrast, our VGrow minimizes the  $f$-divergence from the primal form.
The most related work of GANs to our VGrow is  \cite{johnson18,nitanda2018gradient,wang17}, where functional gradient (first variation of functional)   is adopted
 to help in GAN training.  \cite{nitanda2018gradient} introduced a gradient layer based on first variation  of  generator loss in WGAN \cite{arjovsky17}  to accelerate convergence of training.
 In \cite{wang17},
a deep energy model was trained
along Stein variational gradient \cite{liu2016stein}, which was the projection of the  first  variation of KL divergence in Theorem \ref{th1} onto a reproducing kernel Hilbert space, see Section \ref{pstein} for the  proof.
 \cite{johnson18}  propose a CFG-GAN that  directly minimizes the KL divergence  via functional gradient descent. In their paper, the update direction is  the  gradient of log density ratio multiplied by a positive scaling function. They empirically set this scaling function to be 1 in their numerical study.
Our VGrow is based on the general $f$-divergence, and Theorem \ref{th1} implies  that the update  direction in KL divergence case is indeed  the gradient of log density ratio, and thus the scaling function should be exactly 1.

\section{Experiments}
We evaluated our model on four benchmark datasets including MNIST \cite{lecun98}, FashionMNIST \cite{xiao17}, CIFAR10 \cite{krizhevsky09} and CelebA \cite{liu15}. Four representative $f$-divergences were tested to demonstrate the effectiveness of the general Variational Gradient flow (VGrow) framework for generative learning.

\subsection{Experimental setup}

\textbf{$f$-divergences.} Theoretically, our model works for the whole $f$-divergence family by simply plugging the $f$-function in. Special cases are obtained when specific $f$-divergences are considered. At the population level, when the KL divergence is adopted, our VGrow naturally gives birth to
  CFG-GAN while the adoptation of JS divergence leads to vanilla GAN. As we proved above, GAN with the logD trick corresponds to our newly discovered "logD" divergence which belongs to the f-divergence family.
 Moreover, we consider the Jeffrey divergence to show that our model is  applicable to  other $f$-divergences. 
We name these four cases VGrow-KL, VGrow-JS, VGrow-logD and VGrow-JF.

\textbf{Datasets.} We chose four benchmark datasets which included three small datasets (MNIST, FashionMNIST, CIFAR10) and one large dataset (CelebA) from GAN literature. Both MNIST and FashionMNIST have a training set of 60k examples and a test set of 10k examples as $28\times28\times1$ bilevel images. CIFAR10 has a training set of 50k examples and a test set of 10k examples as $32\times32\times 3$ color images. There are naturally 10 classes on these three datasets. CelebA consists of more than 200k celebrity images which were randomly divided into a training set and a test set by us. The division ratio is approximately 9 : 1. For MNIST and FashionMNIST, the input images were resized to $32\times32\times3$ resolution. We also pre-processed CelebA images by first taking a $160\times160$ central crop and then resizing to the $64\times64\times$ resolution. Only the training sets are used to train our models.

\textbf{Evaluation metrics.} \emph{Inception Score} (IS) \cite{salimans16}, calculates the exponential mutual information $\exp( \mathbb{E}_g {\rm KL} (p(y|g) \Vert p(y)))$ where $p(y|g)$ is the conditional class distribution given the generated image $g$ and $p(y)$ is the marginal class distribution across generated images \cite{barratt18}. To estimate $p(y|g)$ and $p(y)$, we trained specific classifiers on MNIST, FashionMNIST, CIFAR10 following \cite{johnson18} using pre-activation ResNet-18 \cite{he16}. All the IS values were calculated over 50k generated images. \emph{Fr\'echet Inception Distance} (FID) \cite{heusel17} computes the Wasserstein-2 distance by fitting Gaussians on real images $x$ and generated images $g$ after propagated through the Inception-v3 model \cite{szegedy16}, i.e. ${\rm FID}(x, g) = \Vert \mu_x - \mu_g \Vert^2_2 + {\rm Tr}(\Sigma_x + \Sigma_g - 2(\Sigma_x \Sigma_g)^{\frac12})$. Particularly, all the FID scores are reported with respect to the 10k test examples on MNIST, FashionMNIST and CIFAR10 via the tensorflow implementation \url{https://github.com/bioinf-jku/TTUR/blob/master/fid.py}. In a nutshell, higher IS and lower FID are better.

\textbf{Network architectures and hyperparameter settings.} We adopted a new architecture modified from the residual networks used in \cite{miyato18}. The modifications were comprised of reducing the number of batch normalization layers and introducing spectral normalization in the deep sampler / generator. The architecture was shared across the three small datasets and most hyperparameters were shared across different divergences. More residual blocks, upsampling and downsampling are employed on CelebA.
In our experiments, we set the batch size to be 64 and use RMSProp as the SGD optimizer when training neural networks. The learning rate is 0.0001 for both the deep sampler and the deep classifier except for 0.0002 on MNIST for VGrow-JF. Inputs to deep samplers are vectors generated from a $\ell = 128 $ dimensional standard normal distribution on all the datasets. The meta-parameters in our VGrow learning procedure are set to be $s = 0.5$ and $\#$ inner loop $= 20$.

%

The sampler and the classifier are parameterized with residual networks. Each ResNet block has a skip-connection. The skip-connection use upsampling / downsampling of its input and a 1x1 convolution if there is upsampling / downsampling in the residual block. We use the identity mapping as the skip-connection if there is no upsamling / downsampling in the residual block. The upsampling is nearest-neighbor upsampling and the downsampling is achieved with mean pooling. Details concerning the networks are listed in Table \ref{g32}, \ref{d32}, \ref{g64}, \ref{d64} in Appendix B.


\subsection{Results}
Through our experiment, We demonstrate empirically that (1) VGrow is very stable in the training phase, and that (2) VGrow can generate high-fidelity samples that are comparable to real samples both visually and quantitatively. Comparisons with the state-of-the-art GANs suggest the effectiveness of VGrow.

\begin{figure}[ht]
\centering
\subfigure[MNIST]{
\begin{minipage}[t]{3in}
\includegraphics[width=\columnwidth]{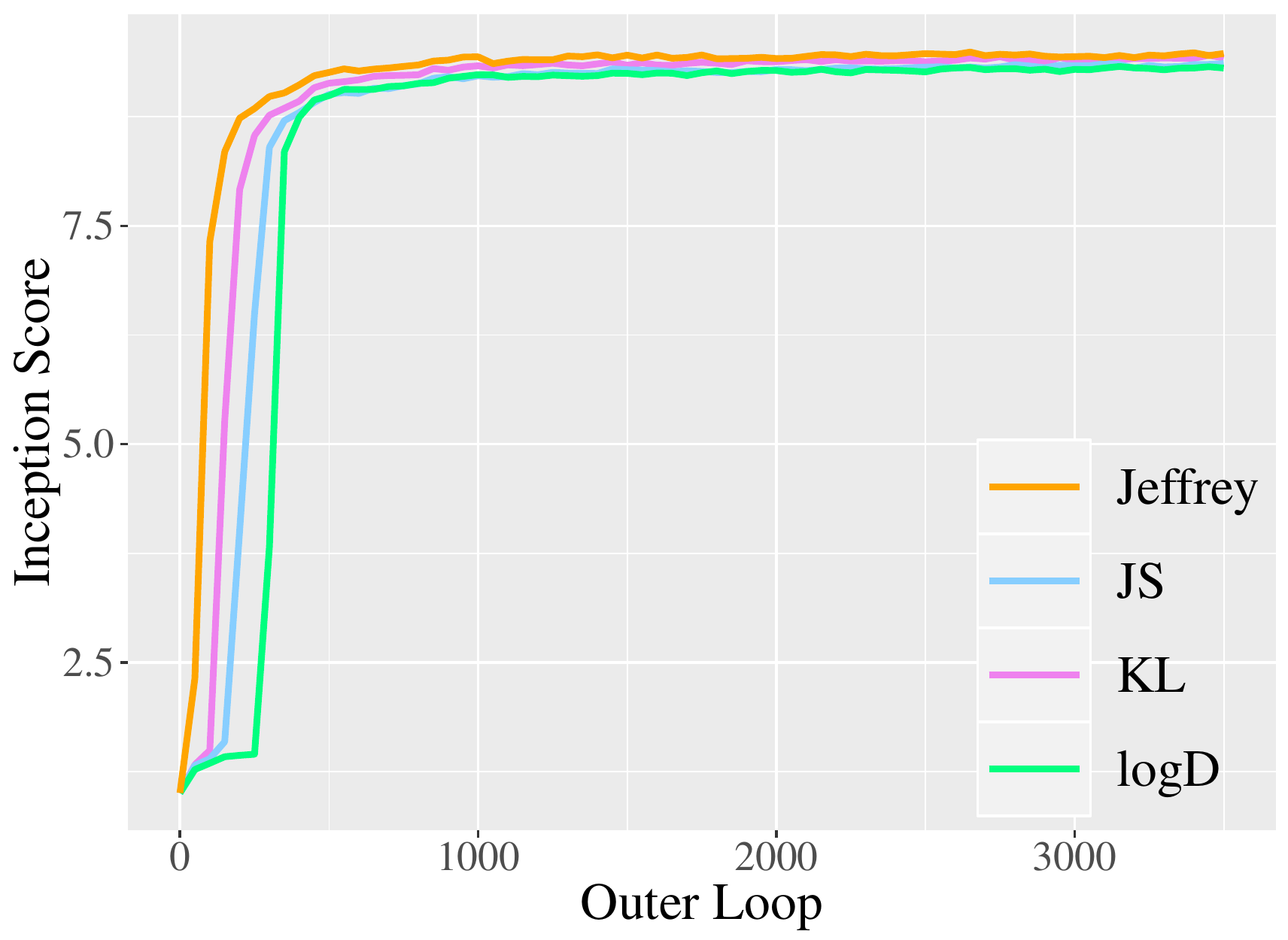}
\end{minipage}
}
\subfigure[FashionMNIST]{
\begin{minipage}[t]{3in}
\includegraphics[width=\columnwidth]{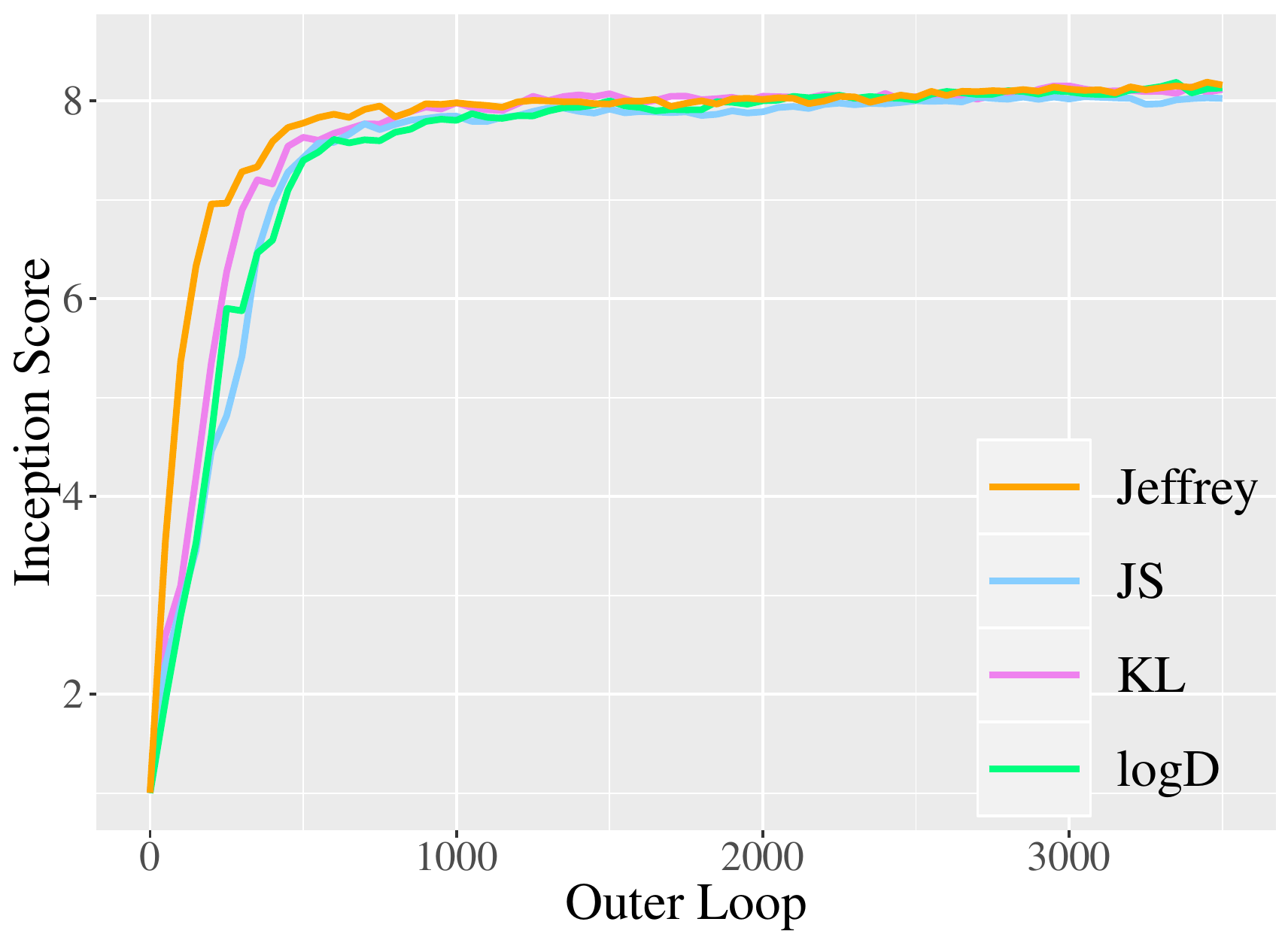}
\end{minipage}
}
\subfigure[CIFAR10]{
\begin{minipage}[t]{3in}
\includegraphics[width=\columnwidth]{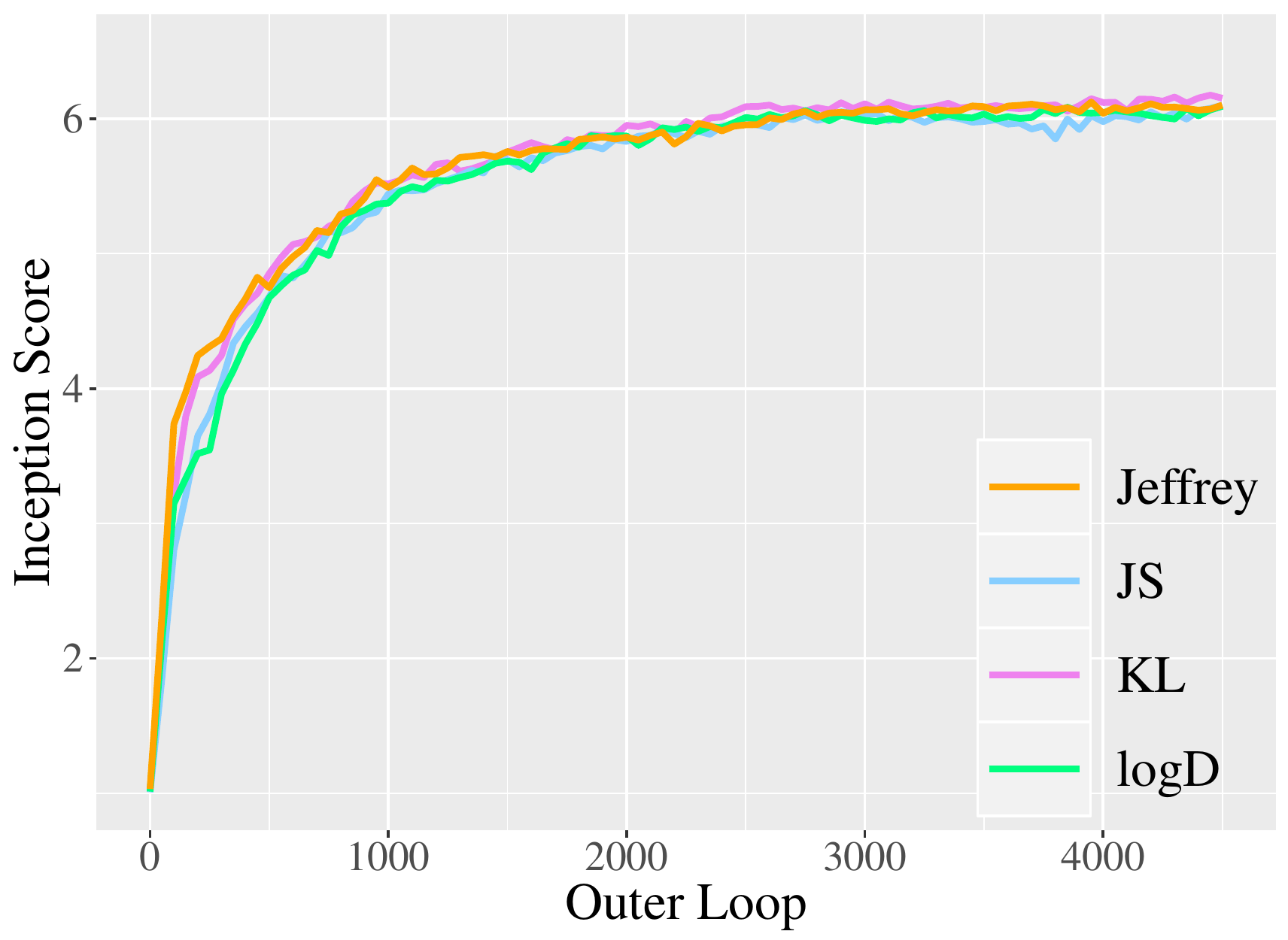}
\end{minipage}
}
\centering
\caption{IS-Loop learning curves on MNIST, FashionMNIST and CIFAR10. The training of VGrow is very stable until 3500 outer loops on MNIST and FashionMNIST (4500 outer loops on CIFAR10).}
\label{IS}
\end{figure}

\textbf{Stability.} It has been shown that the binary classification loss poorly correlates with the generating performance for JS divergence based GAN models \cite{arjovsky17}. We observed similar phenomena with our $f$-divergence based VGrow model, i.e. the classification loss changed a little at the beginning of training and then fluctuated around a constant value. Since the classfication loss was not meaningful enough to measure the generating performance, we turned to utilize the aforementioned inception score to draw IS-Loop \ learning curves on MNIST, FashionMNIST and CIFAR10. The results are presented in Figure \ref{IS}. 
As indicated in all three subfigures, the IS-Loop learning curves are very smooth and the inception scores nearly monotonically increase until 3500 outer loops (almost 75 epochs) on MNIST and FashionMNIST as well as 4500 outer loops (almost 100 epochs) on CIFAR10.

\begin{figure}[ht]
\centering
\subfigure[real MNIST]{
\begin{minipage}[t]{3in}
\includegraphics[width=3in]{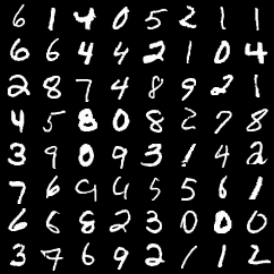}
\end{minipage}
}
\subfigure[generated MNIST]{
\begin{minipage}[t]{3in}
\includegraphics[width=3in]{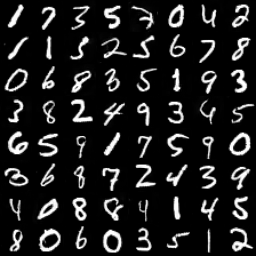}
\end{minipage}
}
\centering
\caption{Real samples and generated samples obtained by VGrow-KL on MNIST.}
\label{samplemnist}
\end{figure}

\begin{figure}
\subfigure[real FashionMNIST]{
\begin{minipage}[t]{3in}
\includegraphics[width=3in]{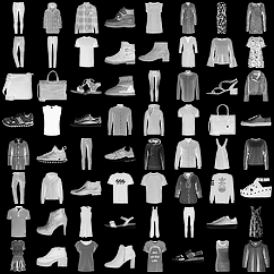}
\end{minipage}
}
\subfigure[generated FashionMNIST]{
\begin{minipage}[t]{3in}
\includegraphics[width=3in]{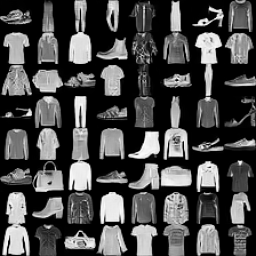}
\end{minipage}
}
\centering
\caption{Real samples and generated samples obtained by VGrow-KL on FashionMNIST.}
\label{samplefmnist}
\end{figure}

\begin{figure}
\subfigure[real CIFAR10]{
\begin{minipage}[t]{3in}
\includegraphics[width=3in]{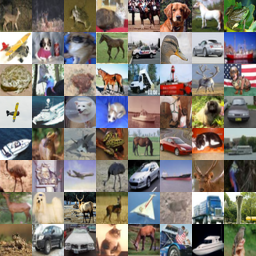}
\end{minipage}
}
\subfigure[generated CIFAR10]{
\begin{minipage}[t]{3in}
\includegraphics[width=3in]{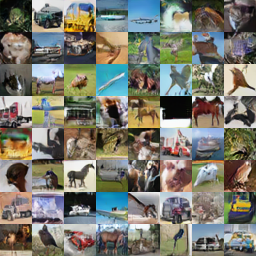}
\end{minipage}
}
\centering
\caption{Real samples and generated samples obtained by VGrow-KL on CIFAR10.}
\label{samplecifar10}
\end{figure}

\begin{figure}
\subfigure[real CelebA]{
\begin{minipage}[t]{3in}
\includegraphics[width=3in]{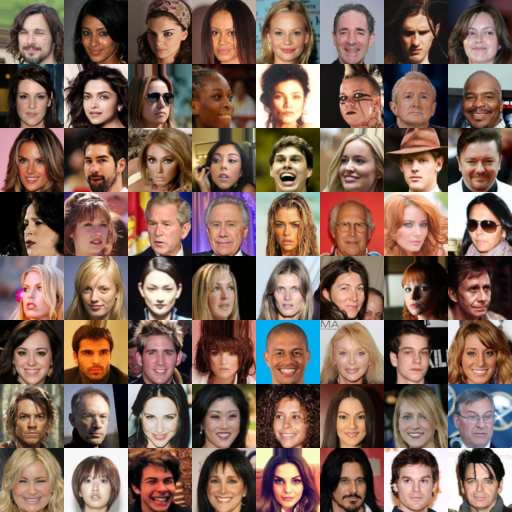}
\end{minipage}
}
\subfigure[generated CelebA]{
\begin{minipage}[t]{3in}
\includegraphics[width=3in]{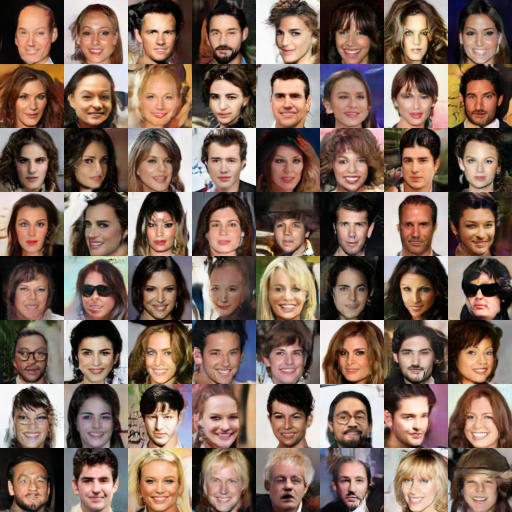}
\end{minipage}
}
\centering
\caption{Real samples and generated samples obtained by VGrow-KL on  CelebA.}
\label{sampleceleba}
\end{figure}

\textbf{Effectiveness.} 
First, we list the real images and generated examples of our VGrow-KL model on the four benchmark datasets in Figure \ref{samplemnist}, \ref{samplefmnist}, \ref{samplecifar10}, \ref{sampleceleba}. We claim that the realistic-looking generated images are visually comparable to real images sampled from the training set. It is easy to distinguish which class the generated example belongs to even on CIFAR10. Second, Table \ref{fid-10k} presents the FID scores for the considered four models, and the FID values on 10k training data of MNIST and FashionMNIST. Scores of generated samples are very close to scores on real data. Especially, VGrow-JS obtains average scores of 3.32 and 8.75 while the scores on training data are 2.12 and 4.16  on MNIST and FashionMNIST, respectively. Third, Table \ref{fid-cf10-50k} shows the FID evaluations of our four models, and the referred evaluations of state-of-the-art WGANs and MMDGANs from \cite{arbel18} based on 50k samples. Our VGrow-logD attain a score of 28.8 with less variance that is competitive with the best (28.5) of referred baseline evalutions. Moreover, VGrow-JS and VGrow-KL achieve better performance than the remaining referred baselines. In a word, the quantitative results in Table \ref{fid-10k} and Table \ref{fid-cf10-50k}  illustrate the effectiveness of our VGrow model.




\begin{table}[t]
\caption{Mean (standard deviation) of FID evaluations over 10k generated MNIST / FashionMNIST images with five-time bootstrap sampling. The last row states statistics of the FID scores between 10k training examples and 10k test examples.}
\label{fid-10k}
\vskip 0.15in
\begin{center}
\begin{small}
\begin{rm}
\begin{tabular}{lcccr}
\toprule
Models 			& MNIST(10k) 			& FashionMNIST (10k) \\
\midrule
VGrow-KL 		& 3.66 (0.09)			& 9.30 (0.09) \\
VGrow-JS 		& \textbf{3.32 (0.05)}	& \textbf{8.75 (0.06)} \\
VGrow-logD 		& 3.64 (0.05)  			& 9.51 (0.09) \\
VGrow-JF 		& 3.40 (0.07)			& 9.72 (0.06) \\
Training set    & 2.12 (0.02)   		& 4.16 (0.03) \\
\bottomrule
\end{tabular}
\end{rm}
\end{small}
\end{center}
\vskip -0.1in
\end{table}
\begin{table}[t]
\caption{Mean (standard deviation) of FID evaluations over 50k generated CIFAR10 images with five-time bootstrap sampling. The last four rows are baseline results adapted from \cite{arbel18}.}
\label{fid-cf10-50k}
\vskip 0.15in
\begin{center}
\begin{small}
\begin{rm}
\begin{tabular}{lcccr}
\toprule
Models 			& CIFAR10 (50k) \\
\midrule
VGrow-KL 		&  29.7 (0.1) \\
VGrow-JS 		&  29.1 (0.1) \\
VGrow-logD 		&  \textbf{28.8 (0.1)} \\
VGrow-JF 		&  32.3 (0.1) \\
\midrule
WGAN-GP			&  31.1 (0.2) \\
MMDGAN-GP-L2	&  31.4 (0.3) \\
SMMDGAN 		&  31.5 (0.4) \\
SN-SWGAN 		&  \textbf{28.5 (0.2)} \\
\bottomrule
\end{tabular}
\end{rm}
\end{small}
\end{center}
\vskip -0.1in
\end{table}
\section{Conclusion}
We propose a framework to learn deep generative models via \textbf{V}ariational \textbf{Gr}adient Fl\textbf{ow} (VGrow) on probability spaces.  We discus connections of our proposed VGrow method with  VAE, GAN and flow-based methods. We  evaluated VGrow on several divergences, including a  newly discovered ``logD'' divergence which serves as the objective function of the logD-trick GAN. Experimental results on benchmark datasets demonstrate that VGrow can generate high-fidelity images in a stable and efficient manner, achieving competitive  performance with state-of-the-art GANs.

\section{Acknowledgment}
The authors would like to thank the reviewers for their
helpful comments. The research presented in this paper
was partially supported by National Science Foundation of  China (NSFC) (Grant number: No.11871474, No.61701547,  No.11501440) and by the research fund of KLATASDS-MOE.

\bibliographystyle{plain}
\bibliography{vgrow}

\clearpage
\section{Appendix A}\label{appa}
In this section we give detail proofs for the main theory in the paper.

\subsection{Proof for Lemma \ref{lem1}}\label{plem1}
\begin{proof}
For any $w(\mathbf{x})$, define the function    $\eta(s) = \mathcal{F} (q +s w): \mathcal{R}^1\rightarrow \mathcal{R}^{1}$.
Chain rule and direct calculation shows     $\eta{'}(s)\big|_{s=0} = \langle\frac{\delta\mathcal{F}}{\delta{q}}(q), w\rangle =  \int f{'}(r(\mathbf{x})) w(\mathbf{x}) d\mathbf{x}$.
\end{proof}

\subsection{Proof for Lemma \ref{lem2}}\label{plem2}
\begin{proof}
Follows from  expression 10.1.16 in \cite{ambrosio2008gradient} (section
E of chapter 10.1.2, page 233.)
\end{proof}

\subsection{Proof for  Theorem \ref{th1}}\label{pth1}
\begin{proof}
For any  $\mathbf{g}(\mathbf{x})\in \mathcal{H}(q_t)$,
define
 $$\eta(s) =  \mathbb{D}_f[({\mathbb{T}_{s, \mathbf{g}}}_\#q_t) (\mathbf{x}) | p(\mathbf{x})]$$
 as a  function of $s\in \mathcal{R}^+$.
Let
$\theta_{\mathbf{g}}(s) = {\mathbb{T}_{s, \mathbf{g}}}_{\#} q_t(\mathbf{x})/p(\mathbf{x}).$
By definition,
$$\mathcal{L}_{}(\mathbf{g}) = \eta(s) = D_f[({\mathbb{T}_{s, \mathbf{g}}}_\#q_t) (\mathbf{x}) | p(\mathbf{x})] = \int p(\mathbf{x}) f(\theta_{\mathbf{g}}(s))\mathrm{d}\mathbf{x}.$$
Since $$\eta{'}(s)\big|_{s=0} = \langle\frac{\delta\mathcal{L}}{\delta \mathbf{h}}(\mathbf{0}), \textbf{g}\rangle_{\mathcal{H}(q_t)},$$ we need calculate the
derivative of $\eta(s)$ at $s=0.$
Recall,
 $$({\mathbb{T}_{s, \mathbf{g}}}_\#q_t) (\mathbf{x}) = q_t(\mathbb{T} ^{-1}_{s, \mathbf{g}}(\mathbf{x})) |\mathrm{det}(\nabla_\mathbf{x} \mathbb{T} ^{-1}_{s, \mathbf{g}}(\mathbf{x}))|,$$
 by chain rule, we get
$$\eta'(s)\big|_{s =0} = \int p(\mathbf{x}) [f'(\theta_{\mathbf{g}}(s)) \theta'_{\mathbf{g}}(s)] \big|_{s=0} \mathrm{d}\mathbf{x},$$
where,
\begin{align*}
&\theta'_{\mathbf{g}}(s)\big|_{s=0} = \frac{1}{p(\mathbf{x})}\Big\{[q_t(\mathbb{T} ^{-1}_{s, \mathbf{g}}(\mathbf{x}))]'\big|_{s=0}|\mathrm{det}(\nabla_\mathbf{x} \mathbb{T} ^{-1}_{s, \mathbf{g}}(\mathbf{x}))|\big|_{s=0}\\
&+q_t(\mathbb{T} ^{-1}_{s, \mathbf{g}}(\mathbf{x}))\big|_{s=0}[|\mathrm{det}(\nabla_\mathbf{x} \mathbb{T} ^{-1}_{s, \mathbf{g}}(\mathbf{x}))|]'\big|_{s=0}\Big\}.
\end{align*}
By definition,
$\theta_{\mathbf{g}}(s)\big|_{s=0} = \frac{q_t(\mathbf{x})}{p(\mathbf{x})}=r_t(\mathbf{x}).$
We claim that
\begin{align*}
\theta'_{\mathbf{g}}(s)\big|_{s=0}
&= \frac{1}{p(\mathbf{x})}\{-\mathbf{g}(\mathbf{x})^T\nabla q_t(\mathbf{x})-q_t(\mathbf{x})\nabla\cdot \mathbf{g}(\mathbf{x})\} \\
&= -\frac{1}{p(\mathbf{x})}{\nabla\cdot[q_t(\mathbf{x})\mathbf{g}(\mathbf{x})]}.
\end{align*}
 Indeed, recall that   $$\mathbb{T}_{s, \mathbf{g}}(X) = X + s \mathbf{g} (X).$$  We get  $$\mathbb{T}_{s, \mathbf{g}}^{-1}(X) = X-s \mathbf{g} (X) + o(s),$$
 and $$\mathbb{T}_{s, \mathbf{g}}^{-1}\big|_{s=0}(X) = {X}.$$  Then it follows that
  \begin{equation*}
  [q_t(\mathbb{T}^{-1}_{s, \mathbf{g}}(\mathbf{x}))]'\big|_{s=0}=\lim\limits_{s \rightarrow0}\frac{q_t(\mathbb{T}^{-1}_{s, \mathbf{g}}(\mathbf{x}))-q_t(\mathbf{x})}{s}= -\mathbf{g}(\mathbf{x})^T\nabla q_t(\mathbf{x}),
   \end{equation*}
   and $$|\mathrm{det}(\nabla_\mathbf{x} \mathbb{T} ^{-1}_{s, \mathbf{g}}(\mathbf{x}))|\big|_{s=0}=1,
q_t(\mathbb{T} ^{-1}_{s, \mathbf{g}}(\mathbf{x}))\big|_{s=0}=q_t(\mathbf{x}).$$

We finish our claim by calculating
\begin{align*}
 &[|\mathrm{det}(\nabla_\mathbf{x} \mathbb{T} ^{-1}_{s, \mathbf{g}}(\mathbf{x}))|]'\big|_{s=0} \\
= & [\exp^{\log(|\mathrm{det}(\nabla_\mathbf{x} \mathbb{T} ^{-1}_{s, \mathbf{g}}(\mathbf{x}))|)}]'\big|_{s=0}\\
= &|\mathrm{det}(\nabla_\mathbf{x} \mathbb{T} ^{-1}_{s, \mathbf{g}}(\mathbf{x}))|\big|_{s=0}[\log |\mathrm{det}(\nabla_\mathbf{x} \mathbb{T} ^{-1}_{s, \mathbf{g}}(\mathbf{x}))|]'|\big|_{s=0}\\
= & \lim\limits_{s\rightarrow 0}\frac{\log |\mathrm{det}(\nabla_\mathbf{x} \mathbb{T} ^{-1}_{s, \mathbf{g}}(\mathbf{x}))|-\log|\mathrm{det}(\mathbf{I})|}{s}\\
= & \lim\limits_{s\rightarrow 0}\frac{\log |\mathrm{det}(\mathbf{I}-s \nabla_\mathbf{x} \mathbf{g}(\mathbf{x}))|-\log|\mathrm{det}(\mathbf{I})|+o(s)}{s}\\
= & -\mathrm{tr}(\nabla_\mathbf{x} \mathbf{g}(\mathbf{x}))= -\nabla\cdot \mathbf{g}(\mathbf{x}).
\end{align*}
Thus,
\begin{align*}
&\eta'_{\mathbf{g}}(s)\big|_{s=0} = \int p(\mathbf{x})\cdot [f'(\theta_{\mathbf{g}}(s))\cdot \theta'_{\mathbf{g}}(s)] \big|_{s=0} \mathrm{d}\mathbf{x}\\
=& -\int f'(r_t(\mathbf{x}))\nabla\cdot[q_t(\mathbf{x})\mathbf{g}(\mathbf{x})] \mathrm{d}\mathbf{x}\\
=& \int  q_t(\mathbf{x})\mathbf{g}(\mathbf{x})^{T}\nabla f'(r_t(\mathbf{x})) - \nabla\cdot[f'(r_t(\mathbf{x}))\mathbf{g}(\mathbf{x})]\mathrm{d}\mathbf{x}\\
=& \int q_t(\mathbf{x})f''(r_t(\mathbf{x}))[\nabla r_t(\mathbf{x})]^T \mathbf{g}(\mathbf{x})\mathrm{d}\mathbf{x} \\
=&  \langle f''(r_t(\mathbf{x}))\nabla r_t(\mathbf{x}), \mathbf{g}(\mathbf{x})\rangle_{\mathcal{H}(q_t)},
\end{align*}
where, the fourth  equality follows from integral  by part and the vanishing assumption.
\end{proof}

\subsection{Proof for Theorem \ref{th2}}\label{pth2}
\begin{proof}
Similar as the proof of equation (13) in \cite{liu2017stein}. We present the detail  here for completeness.
The proof of Theorem \ref{th1} shows that, $$q_{t}(\mathbb{T}^{-1}_{s, \mathbf{v}_t}(\mathbf{x}))=  q_{t}(\mathbf{x}) -s\mathbf{v}_t(\mathbf{x})^T\nabla q_t(\mathbf{x}) +o(s),$$ and $$|\mathrm{det}(\nabla_\mathbf{x} \mathbb{T} ^{-1}_{s, \mathbf{v}_t}(\mathbf{x}))| = -s\nabla\cdot \mathbf{v}_t(\mathbf{x})+o(s).$$
Then by Taylor expansion,
\begin{align*}
& \log ({\mathbb{T}_{s, \mathbf{v}_t}}_\#q_t) (\mathbf{x}) \\
= &\log q_t(\mathbb{T} ^{-1}_{s, \mathbf{v}_t}(\mathbf{x})) +\log  |\mathrm{det}(\nabla_\mathbf{x} \mathbb{T} ^{-1}_{s, \mathbf{v}_t}(\mathbf{x}))| \\
= & \log q_t(\mathbf{x}) - s\frac{\mathbf{v}_t(\mathbf{x})^T\nabla q_t(\mathbf{x})}{ q_t(\mathbf{x})}-s\nabla\cdot \mathbf{v}_t(\mathbf{x})+o(s)\\
= & \log q_t(\mathbf{x}) - \frac{s}{ q_t(\mathbf{x})} ( \mathbf{v}_t(\mathbf{x})^T\nabla q_t(\mathbf{x})\\
  &+  q_t(\mathbf{x})\nabla\cdot \mathbf{v}_t(\mathbf{x}))+o(s).
\end{align*}
 Let $\tilde{q}(\mathbf{x})$ denote the density of ${\mathbb{T}_{s, \mathbf{v}_t}}_\#q_t$.
Then,
\begin{align*}
&\frac{\tilde{q}(\mathbf{x}) - q_t(\mathbf{x})}{s}
=\frac{q_t(\log \tilde{q}-\log q_t)}{s}\\
& = -\nabla\cdot (q_t(\mathbf{x})\mathbf{v}_t(\mathbf{x})) + o(s).
\end{align*}
 Let $s\rightarrow 0$, we get the desired result.
\end{proof}

\subsection{Proof for Lemma \ref{lem3}}\label{plem3}
\begin{proof}
$d^*(\mathbf{x})$ is the minimizer of
\begin{align*}
&\min\limits_{d(\mathbf{x})}\mathbb{E}_{(X, Y) \sim p(\mathbf{x}, y)}\log(1+\exp(-d(\mathbf{X}) Y))\\
=& \min\limits_{d(\mathbf{x})}\int p(\mathbf{x}, y)\log(1+\exp(-d(\mathbf{x}) y))\mathrm{d}\mathbf{x}\mathrm{d}y\\
=& \min\limits_{d(\mathbf{x})}\{\int p(y=1)p(\mathbf{x}|y=1)\log(1+\exp(-d(\mathbf{x})))\mathrm{d}\mathbf{x} \\
& + \int p(y=-1)p(\mathbf{x}|y=-1)\log(1+\exp(d(\mathbf{x})))\mathrm{d}\mathbf{x}\}.
\end{align*}
The above criterion is a functional of $d(\cdot)$. By setting  the first variation to  zero yields
$$\exp^{-d^*(\mathbf{x})}=\dfrac{p(y=1)p(\mathbf{x}|y=1)}{p(y=-1)p(\mathbf{x}|y=-1)},$$
i.e.,
 $r(\mathbf{x}) = \exp^{-d^*(\mathbf{x})}.$
\end{proof}

\subsection{Proof for Theorem \ref{th3}}
\begin{proof}
By definition,
\begin{align*}
&\mathbb{D}_f(q(\mathbf{x}) \Vert p(\mathbf{x}))
= \int p(\mathbf{x}) f\left(\frac{q(\mathbf{x})}{p(\mathbf{x})} \right)d\mathbf{x}\\
=& \int  \left( p(\mathbf{x}) + q(\mathbf{x})\right) \log \left( \frac{p(\mathbf{x}) + q(\mathbf{x})}{p(\mathbf{x})} \right)d\mathbf{x}  - 2 \log2
\end{align*}
At the population level,
the objective function of the logD-trick GAN reads \cite{goodfellow14}:
\begin{align*}
& \max_D    \mathbb{E}_{X \sim p(\mathbf{x})} [\log D(X)] + \mathbb{E}_{Z \sim p_{Z}} [\log (1 - D(G(Z)))],\\
& \min_G - \mathbb{E}_{X \sim p(\mathbf{x})} [\log D(X)]- \mathbb{E}_{Z \sim p_Z} [\log D(G(Z))],
\end{align*}
where, $p_Z$ is the simple low dimensional reference distribution.  Denote   $q(\cdot)$ as  the distribution of $G(Z)$.
Then the losses of $D$ and $G$  are  equivalent to
\begin{align*}
& \max_D    \mathbb{E}_{X \sim p(\mathbf{x})} [\log D(X)] + \mathbb{E}_{X \sim q(\mathbf{x})} [\log (1 - D(X))],\\
& \min_G - \mathbb{E}_{X \sim p(\mathbf{x})} [\log D(X)]- \mathbb{E}_{X\sim q(\mathbf{x})} [\log D(X)].
\end{align*}
The optimal discriminator is $D^*(\mathbf{x}) = \frac{p(\mathbf{x})}{p(\mathbf{x}) + q(\mathbf{x})}.$
Substituting  this $D^*$ into the $G$ criterion,  we get
\begin{align*}
& - \mathbb{E}_{X \sim p(\mathbf{x})} [\log D^*(X)] - \mathbb{E}_{X \sim q(\mathbf{x})} [\log D^*(X)] \\
=& \mathbb{E}_{X \sim p(\mathbf{x})} \left[\log \frac{p(X) + q(X)} {p(X)} \right]  + \mathbb{E}_{X \sim q(\mathbf{x})} \left[ \log \frac{p(X) + q(X)} {p(X)} \right] \\
=& \mathbb{D}_f(q(\mathbf{x}) \Vert p(\mathbf{x})) + 2 \log2. \\
\end{align*}
\end{proof}
\subsection{Proof of the relation of VGrow with SVGD}\label{pstein}
\begin{proof}
Let
$f(u) = u\log u$.
Let $\mathbf{g}$ in a Stein class associate with $q_t$.
By the proof of Theorem \ref{th1}, we know,
\begin{align*}
&\langle\frac{\delta\mathcal{L}}{\delta \mathbf{h}}(\mathbf{0}), \textbf{g}\rangle_{\mathcal{H}(q_t)} \\
=&  \langle f''(r_t)\nabla r_t, \textbf{g}\rangle_{\mathcal{H}(q_t)}\\
=&\int \mathbf{g}(\mathbf{x})^T\frac{\nabla r_t(\mathbf{x})}{r_t(\mathbf{x})}q_t(\mathbf{x})\mathrm{d}\mathbf{x}\\
=&\int \mathbf{g}(\mathbf{x})^T\nabla \log r_t(\mathbf{x})q_t(\mathbf{x})\mathrm{d}\mathbf{x}\\
=&\mathbb{E}_{X\sim q_t(\mathbf{x})}[ \mathbf{g}(\mathbf{x})^T\nabla \log q_t(X) -  \mathbf{g}(\mathbf{x})^T\nabla \log p(X)]\\
=&\mathbb{E}_{X\sim q_t(\mathbf{x})}[\mathbf{g}(\mathbf{x})^T\nabla \log q_t(X) + \nabla\cdot \mathbf{g}(\mathbf{x})]\\
 &- \mathbb{E}_{X\sim q_t(\mathbf{x})}[\mathbf{g}(\mathbf{x})^T\nabla \log p(X) + \nabla\cdot \mathbf{g}(\mathbf{x})]\\
=&\mathbb{E}_{X\sim q_t(\mathbf{x})}[\mathcal{T}_{q_t} \mathbf{g}]- \mathbb{E}_{X\sim q_t(\mathbf{x})}[\mathcal{T}_{p} \mathbf{g}]\\
=& - \mathbb{E}_{X\sim q_t(\mathbf{x})}[\mathcal{T}_{p} \mathbf{g}],
\end{align*}
where, last equality follows from via restricting $\mathbf{g}$ in a Stein class associate with $q_t$, i.e., $\mathbb{E}_{X\sim q_t(\mathbf{x})}\mathcal{T}_{q_t} \mathbf{g} = 0$.
\end{proof}

\section{Appendix B}\label{appb}
In this Section, We present the detail of the network used in our experiment.
We use $c$ to denote the number of channels of the images used in the experiment, i.e., $c=1$ or $c=3.$
\begin{table}[ht]
\caption{ResNet sampler with $32 \times 32\times c $ resolution. }
\label{g32}
\vskip 0.15in
\begin{center}
\begin{small}
\begin{rm}
\begin{tabular}{lcccr}
\toprule
Layer    & Details & Output size \\
\midrule
Latent noise 	& $\bm{z} \sim \mathcal{N} (0, I)$	 & 128  \\
\midrule
Fully connected & Linear 						& 2048 \\
				& Reshape 						& $4 \times 4 \times 128$ \\
\midrule
ResNet block 	& ReLU 							& $4 \times 4 \times 128$ \\
				& Upsampling 						& $8 \times 8 \times 128$ \\
				& Conv$3 \times 3$, BN, ReLU 	& $8 \times 8 \times 128$ \\
				& Conv$3 \times 3$ 				& $8 \times 8 \times 128$ \\
\midrule
ResNet block 	& ReLU 							& $8 \times 8 \times 128$ \\
				& Upsampling 						& $16 \times 16 \times 128$ \\
				& Conv$3 \times 3$, BN, ReLU 	& $16 \times 16 \times 128$ \\
				& Conv$3 \times 3$ 				& $16 \times 16 \times 128$ \\
\midrule
ResNet block 	& ReLU 							& $16 \times 16 \times 128$ \\
				& Upsampling 						& $32 \times 32 \times 128$ \\
				& Conv$3 \times 3$, BN, ReLU 	& $32 \times 32 \times 128$ \\
				& Conv$3 \times 3$ 				& $32 \times 32 \times 128$ \\
\midrule
Conv			& ReLU, Conv$3 \times 3$, Tanh	& $32 \times 32 \times c$ \\
\bottomrule
\end{tabular}
\end{rm}
\end{small}
\end{center}
\vskip -0.1in
\end{table}

\begin{table}[ht]
\caption{ResNet classifier with $32 \times 32\times c$ resolution.}
\label{d32}
\vskip 0.15in
\begin{center}
\begin{small}
\begin{rm}
\begin{tabular}{lcccr}
\toprule
Layer    		& Details 						& Output size \\
\midrule
ResNet block 	& Conv$3 \times 3$		 		& $32 \times 32 \times 128$ \\
				& ReLU, Conv$3 \times 3$ 		& $32 \times 32 \times 128$ \\
				& Downsampling 					& $16 \times 16 \times 128$ \\
\midrule
ResNet block 	& ReLU, Conv$3 \times 3$ 		& $16 \times 16 \times 128$ \\
				& ReLU, Conv$3 \times 3$ 		& $16 \times 16 \times 128$ \\
				& Downsampling 					& $8 \times 8 \times 128$ \\
\midrule
ResNet block 	& ReLU, Conv$3 \times 3$ 		& $8 \times 8 \times 128$ \\
				& ReLU, Conv$3 \times 3$ 		& $8 \times 8 \times 128$ \\
\midrule
ResNet block 	& ReLU, Conv$3 \times 3$ 		& $8 \times 8 \times 128$ \\
				& ReLU, Conv$3 \times 3$ 		& $8 \times 8 \times 128$ \\
\midrule
Fully connected & ReLU, GlobalSum pooling 		& 128 \\
				& Linear 						& 1 \\
\bottomrule
\end{tabular}
\end{rm}
\end{small}
\end{center}
\vskip -0.1in
\end{table}

\begin{table}[ht]
\caption{ResNet sampler with $64 \times 64\times c$ resolution. }
\label{g64}
\vskip 0.15in
\begin{center}
\begin{small}
\begin{rm}
\begin{tabular}{lcccr}
\toprule
Layer    & Details & Output size \\
\midrule
Latent noise 	& $\bm{z} \sim \mathcal{N} (0, I)$	 & 128  \\
\midrule
Fully connected & Linear 						& 2048 \\
				& Reshape 						& $4 \times 4 \times 128$ \\
\midrule
ResNet block 	& ReLU 							& $4 \times 4 \times 128$ \\
				& Upsampling 					& $8 \times 8 \times 128$ \\
				& Conv$3 \times 3$, BN, ReLU 	& $8 \times 8 \times 128$ \\
				& Conv$3 \times 3$ 				& $8 \times 8 \times 128$ \\
\midrule
ResNet block 	& ReLU 							& $8 \times 8 \times 128$ \\
				& Upsampling 					& $16 \times 16 \times 128$ \\
				& Conv$3 \times 3$, BN, ReLU 	& $16 \times 16 \times 128$ \\
				& Conv$3 \times 3$ 				& $16 \times 16 \times 128$ \\
\midrule
ResNet block 	& ReLU 							& $16 \times 16 \times 128$ \\
				& Upsampling 					& $32 \times 32 \times 128$ \\
				& Conv$3 \times 3$, BN, ReLU 	& $32 \times 32 \times 128$ \\
				& Conv$3 \times 3$ 				& $32 \times 32 \times 128$ \\
\midrule
ResNet block 	& ReLU 							& $32 \times 32 \times 128$ \\
				& Upsampling 					& $32 \times 32 \times 128$ \\
				& Conv$3 \times 3$, BN, ReLU 	& $32 \times 32 \times 128$ \\
				& Conv$3 \times 3$ 				& $64 \times 64 \times 128$ \\
\midrule
Conv			& ReLU, Conv$3 \times 3$, Tanh	& $64 \times 64 \times c$ \\
\bottomrule
\end{tabular}
\end{rm}
\end{small}
\end{center}
\vskip -0.1in
\end{table}

\begin{table}[ht]
\caption{ResNet classifier with $64 \times 64\times c$ resolution. }
\label{d64}
\vskip 0.15in
\begin{center}
\begin{small}
\begin{rm}
\begin{tabular}{lcccr}
\toprule
Layer    		& Details 						& Output size \\
\midrule
ResNet block 	& Conv$3 \times 3$				& $64 \times 64 \times 128$ \\
				& ReLU, Conv$3 \times 3$ 		& $64 \times 64 \times 128$ \\
				& Downsampling 					& $32 \times 32 \times 128$ \\
\midrule
ResNet block 	& ReLU, Conv$3 \times 3$		& $32 \times 32 \times 128$ \\
				& ReLU, Conv$3 \times 3$ 		& $32 \times 32 \times 128$ \\
				& Downsampling 					& $16 \times 16 \times 128$ \\
\midrule
ResNet block 	& ReLU, Conv$3 \times 3$ 		& $16 \times 16 \times 128$ \\
				& ReLU, Conv$3 \times 3$ 		& $16 \times 16 \times 128$ \\
				& Downsampling 					& $8 \times 8 \times 128$ \\
\midrule
ResNet block 	& ReLU, Conv$3 \times 3$ 		& $8 \times 8 \times 128$ \\
				& ReLU, Conv$3 \times 3$ 		& $8 \times 8 \times 128$ \\
\midrule
Fully connected & ReLU, GlobalSum pooling 		& 128 \\
				& Linear 						& 1 \\
\bottomrule
\end{tabular}
\end{rm}
\end{small}
\end{center}
\vskip -0.1in
\end{table}

\end{document}